\pdfoutput=1

\documentclass[11pt]{article}

\usepackage[]{emnlp2021}

\usepackage{times}
\usepackage{latexsym}
\usepackage{empheq}
\usepackage{bm}

\usepackage{booktabs}       
\usepackage{amsfonts}       
\usepackage{nicefrac}       
\usepackage{todonotes}
\usepackage{paralist}

\usepackage[T1]{fontenc}

\usepackage[utf8]{inputenc}

\usepackage{microtype, todonotes}
\usepackage{array}
\newcolumntype{x}[1]{>{\centering\let\newline\\\arraybackslash\hspace{0pt}}p{#1}}

\title{Uncertainty Measures in Neural Belief Tracking and\\ the Effects on Dialogue Policy Performance}

\author{Carel van Niekerk$^1$, Andrey Malinin$^{2}$, Christian Geishauser$^1$, Michael Heck$^1$ \\
{\bf Hsien-chin Lin$^1$, Nurul Lubis$^1$, Shutong Feng$^1$ and Milica Ga\v{s}i\'c$^1$} \\
  $^1$Heinrich Heine Universität Düsseldorf, Düsseldorf, Germany \\
  $^2$Yandex Research and HSE University, Moscow, Russia \\
  $^1$\texttt{\{niekerk,geishaus,heckmi,linh,lubis,shutong.feng,gasic\}@hhu.de} \\
  $^2$\texttt{am969@yandex-team.ru} \\}

\usepackage{xcolor,colortbl}
\usepackage{multirow}

\definecolor{Gray}{gray}{0.85}
\definecolor{LightCyan}{rgb}{0.88,1,1}

\newcolumntype{a}{>{\columncolor{Gray}}c}
\newcolumntype{b}{>{\columncolor{white}}c}

\begin{document}

\maketitle

\begin{abstract}
The ability to identify and resolve uncertainty is crucial for the robustness of a dialogue system. Indeed, this has been confirmed empirically on systems that utilise Bayesian approaches to dialogue belief tracking.  However, such systems consider only confidence estimates and have difficulty scaling to more complex settings.  Neural dialogue systems, on the other hand, rarely take uncertainties into account. They are therefore overconfident in their decisions and less robust.  Moreover, the performance of the tracking task is often evaluated in isolation, without consideration of its effect on the downstream policy optimisation.  We propose the use of different uncertainty measures in neural belief tracking.  The effects of these measures on the downstream task of policy optimisation are evaluated by adding selected measures of uncertainty to the feature space of the policy and training policies through interaction with a user simulator. Both human and simulated user results show that incorporating these measures leads to improvements both of the performance and of the robustness of the downstream dialogue policy. This highlights the importance of developing neural dialogue belief trackers that take uncertainty into account.
\end{abstract}

\section{Introduction}

In task-oriented dialogue, the system aims to assist the user in obtaining information. This is achieved through a series of interactions between the user and the system. As the conversation progresses, it is the role of the \emph{dialogue state tracking} module to track the state of the conversation. For example, in a restaurant recommendation system, the state would include the information about the cuisine of the desired restaurant, its area as well as the price range that the user has in mind. It is crucial that this state contains all information necessary for the \emph{dialogue policy} to make an informed decision for the next action~\citep{young2007his}. Policy training optimises decision making in order to complete dialogues successfully.

It has been proposed within the partially observable Markov decision process (POMDP) approach to dialogue modelling to track the distribution over all possible dialogue states, the \emph{belief state}, instead of a single most-likely candidate. This approach successfully integrates uncertainty to achieve robustness~\citep{williams2007pomdp, thomson2010bayesian, young2016evaluation,young2007his}. However, such systems do not scale well to complex multi-domain dialogues. On the other hand, discriminative neural approaches to dialogue tracking achieve state-of-the-art performance in the state tracking task. Nevertheless, the state-of-the-art goal accuracy on the popular MultiWOZ~\citep{budzianowski2018large} multi-domain benchmark is currently only at $60\%$~\citep{heck2020trippy, li2020coco}. In other words, even the best neural dialogue state trackers at present incorrectly predict the state of the conversation in $40\%$ of the turns. What is particularly problematic is that these models are fully confident about their incorrect predictions.

Unlike neural dialogue state trackers, which predict a single best dialogue state, neural belief trackers produce a belief state~\citep{williams2007pomdp, henderson-etal-2013-deep}. State-of-the-art neural belief trackers, however, achieve an even lower goal accuracy of approximately $50\%$ \citep{van-niekerk-2020-knowing,lee2019sumbt}, making the more accurate state trackers a preferred approach. High-performing state trackers typically rely on span-prediction approaches, which are unable to produce a distribution over all possible states as they extract information directly from the dialogue.

Ensembles of models are known to yield improved predictive performance as well as a calibrated and rich set of uncertainty estimates~\cite{malinin-thesis, galthesis}.
Unfortunately, ensemble generation and, especially, inference come at a high computational and memory cost which may be prohibitive. While standard ensemble distillation~\citep{hinton2015distilling} can be used to compress an ensemble into a single model, information about ensemble diversity, and therefore several uncertainty measures, is lost. Recently \citet{malinin2019ensemble} and \citet{malinin2021send} proposed \emph{ensemble distribution distillation} (EnD$^2$) - an approach to distill an ensemble into a single model which preserves both the ensemble's improved performance and full set of uncertainty measures at low inference cost.

In this work we use EnD$^2$ to distill an ensemble of neural belief trackers into a single model and incorporate additional uncertainty measures, namely confidence scores, total uncertainty (entropy) and knowledge uncertainty (mutual information), into the belief state of the neural dialogue system. This yields an uncertainty-aware neural belief tracker and allows downstream dialogue policy models to use this information to resolve confusion. To our knowledge, ensemble distillation, especially ensemble distribution distillation, and the derived uncertainty estimates, have not been examined for belief state estimation or \emph{any} downstream tasks.

We make the following contributions:

\begin{compactenum}
    \item We present SetSUMBT, a modified SUMBT belief tracking model, which incorporates set similarity for accurate state predictions and produces components essential for policy optimisation.

    \item We deploy ensemble distribution distillation to obtain well-calibrated, rich estimates of uncertainty in the dialogue belief tracker. The resulting model produces state-of-the-art results in terms of calibration measures.

    \item We demonstrate the effect of adding additional uncertainty measures in the belief state on the downstream dialogue policy models and confirm the effectiveness of these measures both in a simulated environment and in a human trial. 
\end{compactenum}

\section{Background}

\subsection{Dialogue Belief Tracking}

In statistical approaches to dialogue, one can view the dialogue as a Markov decision process (MDP)~\citep{levin1998mdp}. This MDP maintains a Markov \emph{dialogue state} in each turn and chooses its next \emph{action} based on this state. 

Alternatively, we can model the dialogue state as a latent variable, maintaining a \emph{belief state} at each turn, as in partially observable Markov decision processes (POMDPs)~\citep{williams2007pomdp, thomson2010bayesian}. While attractive in theory, the POMDP model is computationally expensive in practice. Although there are practical implementations, they are limited to single-domain dialogues and their performance fall short of discriminative statistical belief trackers~\citep{williams-2012-belief}. The inherent problem lies in the generative nature of POMDP trackers where the state generates noisy observations. This becomes an issue for instance when the user wants to change the goal of a conversation, e.g., the user wants an Italian instead of a French restaurant. \citet{henderson2015discriminative} has shown empirically that discriminative models model a change in user goal more accurately. 

In discriminative approaches, the state depends on the observation, making it easier for the system to identify a change of the user goal. Traditional discriminative approaches suffer from low robustness, as they depend on static semantic dictionaries for feature extraction~\citep{henderson2014robust, mrkvsic2016nbt}. Integrated approaches on the other hand utilise learned token vector representations, leading to more robust state trackers~\citep{mrksic-etal-2017-neural,ramadan2018mdbt,lee2019sumbt}. However, highly over-parameterised models, such as neural networks -- when trained via maximum-likelihood on finite data -- often yield miscalibrated, over-confident predictions, placing \emph{all} probability mass on a single outcome~\cite{calibration2017}. Consequently, belief tracking is reduced to state tracking, losing the benefits of uncertainty management. State-of-the-art approaches to dialogue state tracking redefine the problem as a span-prediction task. These models extract the values directly from the dialogue context~\citep{chao2019bertdst, zhang2019dsdst, heck2020trippy} and manage to achieve state-of-the-art results on MultiWOZ~\citep{budzianowski2018large, eric2019multiwoz}. Span-prediction models at present do not produce probability distributions, so additional work is needed to apply our proposed uncertainty framework to them. Neural belief and state trackers rarely model the correlation between domain-slot pairs, except for works by \citet{hu2020sas} and \citet{ye2021slot}. Due to scalability issues we do not include these approaches in our investigation. We therefore consider the slot-utterance matching belief tracker (SUMBT)~\citep{lee2019sumbt} a better starting point, as it is readily able to produce a belief state distribution.

In theory, well-calibrated belief trackers have an inherent advantage over state tracking, producing uncertainty estimates that lead to more robust downstream policy performance. This raises the question: Is it possible to instil well-calibrated uncertainty estimates in neural belief trackers? And if so, do these estimates have a positive effect on the downstream policy optimisation in practice?


We believe SUMBT is a fitting approach to investigate these questions, as it has been shown that an ensemble of SUMBT models can achieve state-of-the-art goal L2-Error when trained using specialised loss functions aiming at inducing uncertainty in the output~\citep{van-niekerk-2020-knowing}.

\subsection{Ensemble-based Uncertainty Estimation}

Consider a classification problem with a set of features $\bm{x}$, and outcomes $y \in \{ \omega_1, \omega_2, ..., \omega_K \}$. In dialogue state tracking, $\bm{x}$ would be features of the input to the tracker and $y$ would be a dialogue state. Given an ensemble of $M$ models $\big\{{\tt P}(y| \bm{x}, \bm{\theta}^{(m)} )\big\}_{m=1}^M$, the \emph{predictive posterior} is obtained as follows:
\begin{empheq}{align}
\begin{split}
{\tt P}(y|\bm{x}, \mathcal{D}) \small{=}& 
                               \sum_{m=1}^M \frac{{\tt P}(y| \bm{x}, \bm{\theta}^{(m)})}{M} \small{\eqqcolon}\sum_{m=1}^M \frac{\bm{\pi}^{(m)}}{M}
\end{split}
\label{eqn:predposterior}
\end{empheq}
Predictions made using the predictive posterior are often better than those of individual models. The entropy $\mathcal{H}[]$ of the predictive posterior is an estimate of \emph{total uncertainty}. Ensembles allow decomposing \emph{total uncertainty} into \emph{data} and \emph{knowledge uncertainty} by considering measures of \emph{ensemble diversity}. \emph{Data uncertainty} is the uncertainty due to noise, ambiguity and class overlap in the data. \emph{Knowledge uncertainty} is uncertainty due to a lack of knowledge of the model about a \emph{test data}~\cite{malinin-thesis,galthesis} --- ie, uncertainty due to unfamiliar, anomalous or atypical inputs. Ideally, ensembles should yield \emph{consistent} predictions on data similar to the training data and \emph{diverse} predictions on data which is significantly different from the training data. Thus measures of ensemble diversity yield estimates of \emph{knowledge uncertainty}\footnote{In-depth overviews of ensemble methods are available in~\citet{malinin-thesis,gal2016dropout,ashukha2020pitfalls,trust-uncertainty}.}. These quantities are obtained via the mutual information $\mathcal{I}[y,\bm{\theta}]$ between predictions and model parameters. The quantity in the Equation~\ref{eqn:mibayes} is a measure of \emph{ensemble diversity}, and therefore, \emph{knowledge uncertainty}. This quantity is the difference between the entropy of the predictive posterior (total uncertainty) and the average entropy of each model in the ensemble (data uncertainty).
\begin{empheq}{align}
\begin{split}
& \underbrace{\mathcal{I}[y,\bm{\theta}| \bm{x},\mathcal{D}]}_{\text{Knowledge unc.}} \small{=} \\ & \underbrace{\mathcal{H}\big[{\tt P}(y|\bm{x}, \mathcal{D})\big]}_{\text{Total uncertainty}}
  \small{-} \underbrace{\sum_{m=1}^M\frac{\mathcal{H}\big[{\tt P}(y|\bm{x},\bm{\theta}^{(m)})\big]}{M}}_{\text{Data uncertainty}} 
\end{split}
\label{eqn:mibayes}
\end{empheq}

\subsection{Ensemble Distillation}\label{sec:distillation}

While ensembles provide improved predictive performance and a rich set of uncertainty measures, their practical application is limited by their inference-time computational cost. Ensemble distillation (EnD)~\cite{hinton2015distilling} can be used to compress an ensemble into a single student model (with parameters $\bm{\phi}$) by minimising the Kullback-Leibler (KL) divergence between the ensemble predictive posterior and the distilled model predictive posterior, significantly reducing the inference cost. Unfortunately, a significant drawback of this method is that information about \emph{ensemble diversity}, and therefore \emph{knowledge uncertainty}, is lost in the process. Recently,~\citet{malinin2019ensemble} proposed \emph{ensemble distribution distillation} (EnD$^2$) as an approach to distill an ensemble into a single \emph{prior network} model~\cite{malinin2018prior}, such that the model retains information about ensemble diversity. Prior networks yield a higher-order \emph{Dirichlet distribution} over categorical output distributions $\bm{\pi}$ and thereby \emph{emulate} ensembles, whose output distributions can be seen as samples from a higher-order distribution\footnote{$\bm{\pi} = [{\tt P}(y =\omega_1 | \bm{x}),\cdots,{\tt P}(y =\omega_K | \bm{x})]^{\tt T}$.}. Formally, a prior network is defined as follows:
\begin{empheq}{align}
\begin{split}
& {\tt p}(\bm{\pi} | \bm{x};\bm{\phi}) = {\tt Dir}(\bm{\pi} | \bm{ \alpha}), \  \bm{ \alpha} = e^{\bm{z}}  \\ & \bm{z} = \bm{f}(\bm{x};\bm{\phi}),\
\alpha_k > 0,\  \alpha_0 = \sum_{k=1}^K  \alpha_k,
\end{split}
\label{eqn:DPN1}
\end{empheq}
where ${\tt Dir}(\cdot | \bm{ \alpha})$ is a Dirichlet distribution with concentration parameters $\bm{\alpha}$, and $\bm{f}(\cdot;\bm{\phi})$ is a learned function which yields the logits $\bm{z}$. The predictive posterior can be obtained in closed form though marginalisation over $\bm{\pi}$, thereby emulating \eqref{eqn:predposterior}. This yields a softmax output function:
\begin{empheq}{align}
\begin{split} 
{\tt P}(y = \omega_k | \bm{x};\bm{ \phi}) = &\ \mathbb{E}_{{\tt p}(\bm{\pi} | \bm{x};\bm{\phi})}[{\tt P}(y = \omega_k | \bm{\pi})] \\
= &\ \frac{ e^{ z_k}}{\sum_{k=1}^K e^{ z_k}}.
\end{split}\label{eqn:dirposterior}
\end{empheq}

\emph{Closed form} estimates of all uncertainty measures are obtained via  Eq.~\eqref{eqn:mipn}, which emulates the same underlying mechanics as Eq.~\eqref{eqn:mibayes}, as follows~\cite{malinin-thesis}:
\begin{empheq}{align}
\begin{split} 
&\underbrace{\mathcal{I}[y,{\tt \bm{\pi}} |\bm{x};\bm{\phi}]}_{\text{Knowledge unc.}}\small{=}\\&\underbrace{\mathcal{H}\big[{\tt P}(y | \bm{x};\bm{ \phi})\big]}_{\text{Total uncertainty}} 
\small{-} \underbrace{\mathbb{E}_{{\tt p}({\tt \bm{\pi}}|\bm{x};\bm{\phi})}\big[\mathcal{H}[{\tt P}(y|{\tt \bm{\pi}})]\big]}_{\text{Data uncertainty}} .
\end{split}\label{eqn:mipn}
\end{empheq}

Originally,~\citet{malinin2019ensemble} implemented EnD$^2$ on the CIFAR10, CIFAR100 and TinyImageNet datasets. However, \citet{malinin2021send} found scaling to tasks with many classes challenging using the original Dirichlet Negative log-likelihood criterion. They analysed this scaling problem and proposed to a new loss function, which minimises the \emph{reverse KL-divergence} between the model and an intermediate \emph{proxy Dirichlet} target derived from the ensemble. This loss function was shown to enable EnD$^2$ on tasks with arbitrary numbers of classes. In this work we use this improved loss function, as detailed in the Appendix Section~\ref{app:end2loss}.

\subsection{Policy Optimisation}
In each turn of dialogue, the dialogue policy selects an action to take in order to successfully complete the dialogue. The input to the policy is constructed using the output of the belief state tracker, thus being directly impacted by its richness. 

Optimising dialogue policies within the original POMDP framework is not practical for most cases. Therefore, the POMDP is viewed as a continuous MDP whose state space is the belief space. This state space can be discretised, so that tabular reinforcement learning (RL) algorithms can be applied~\cite{gasic08, thomson10}. Gaussian process RL can be applied directly on the original belief space~\cite{gasic14}. This is also possible using neural approaches with less computational effort~\cite{beliefcritic, acer, strac}. Current state-of-the-art RL algorithms for multi-domain dialogue management~\cite{gdpl, gdplwo} utilise proximal policy optimisation~\cite{ppo} operating on single best dialogue state.

\section{Effects of Uncertainty on Downstream Tasks}

We take the following steps in order to examine the effects of the additional uncertainty measures in the dialogue belief state:
\begin{compactenum}
    \item Modify the original SUMBT model~\cite{lee2019sumbt} to arrive at a competitive baseline. We call this model SetSUMBT.
    \item Produce ensembles of SetSUMBT following the work
    of~\citet{van-niekerk-2020-knowing}.
    \item Apply EnD and EnD$^2$ as introduced in~Section~\ref{sec:distillation}.
    \item Apply policy optimisation that uses belief states from distilled models.
\end{compactenum}

\subsection{Neural Belief Tracking Model}
\begin{figure*}[t]
    \centering
    \includegraphics[scale=0.45]{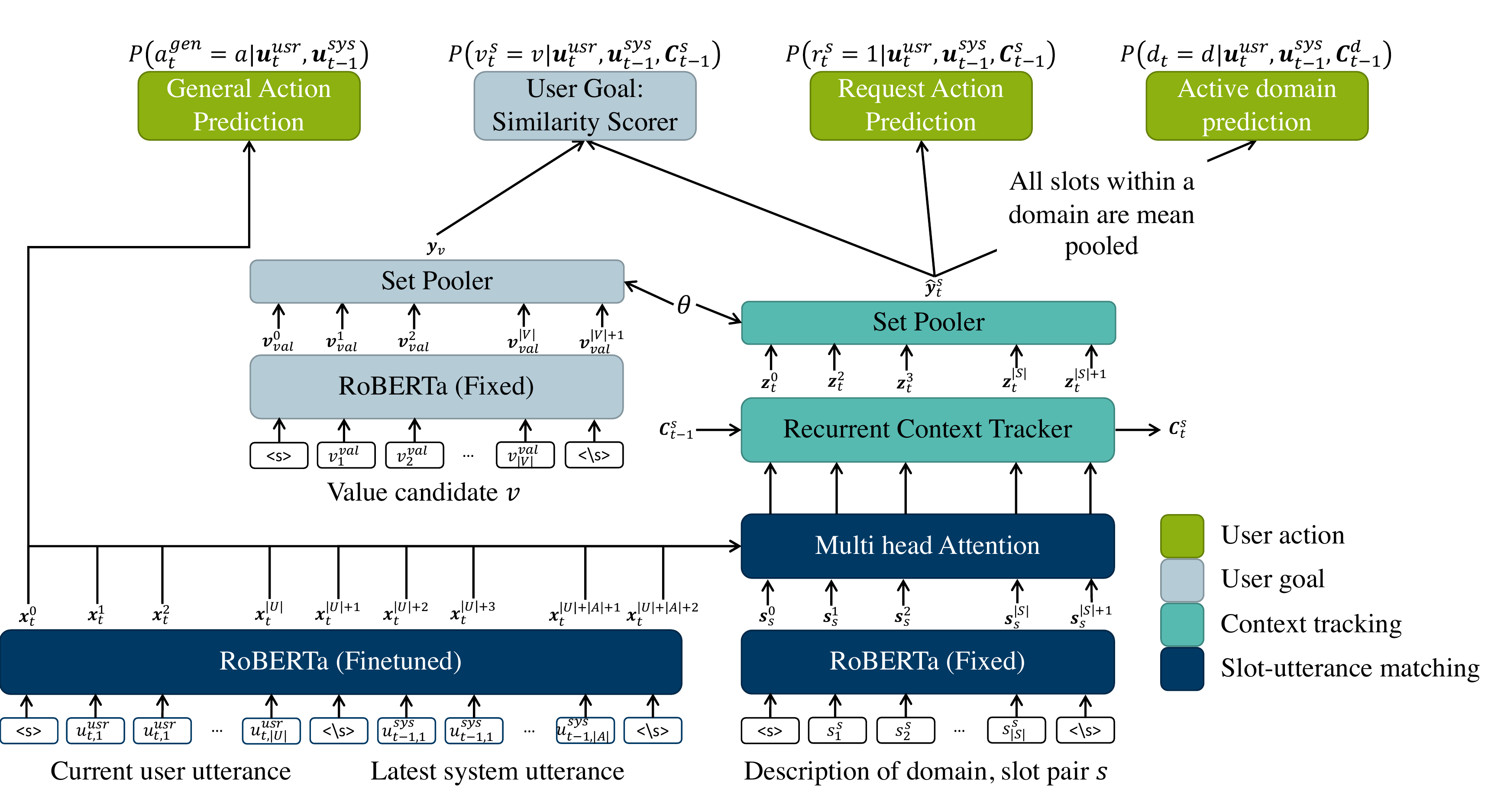}
    \caption{Architecture of our SetSUMBT model, which takes as input the current user utterance, the latest system utterance, and a domain-slot pair description. The model, further, requires a pre-defined set of plausible value candidates for each domain-slot pair. At each turn, we encode the utterances only once, the \emph{Slot-utterance matching} and \emph{Context tracking} components are utilised once for each domain-slot pair. Further, we use the \emph{Set Pooler} once for each domain-slot pair and once for each value candidate. The \emph{Set Pooler} used for pooling value candidate and domain-slot context sequences shares the same parameters $\bm{\theta}$. SetSUMBT outputs a belief state distribution for the relevant domain-slot pair (\emph{User goal}), a distribution over general actions, and the probability of a user request for the domain-slot pair (\emph{User action}). The model also outputs the probability of an active domain.}
    \label{fig:setsumbt}
\end{figure*}

We propose a neural belief tracker which one can easily incorporate in a full dialogue system pipeline. We base our tracker on the slot-utterance matching belief tracker (SUMBT)~\citep{lee2019sumbt}, but we make two important changes. First, we ensure our tracker is fully in line with the requirements of the hidden information state (HIS) model for dialogue management~\citep{young2007his} by adding user action predictions to our tracker. These are not produced by the SUMBT model and nor by other available neural trackers. However, they are essential for integration into a full dialogue system. Second, in order to improve the understanding ability of the model, we utilise a set of concept description embeddings rather than a single embedding for semantic concepts. We use this set of embeddings for information extraction and prediction, hence we call our model SetSUMBT. In this section we describe each component in detail, also depicted in Figure~\ref{fig:setsumbt}.

\paragraph{Slot-utterance matching} The slot-utterance matching (SUM) component performs the role of language understanding in the SUMBT architecture. The SUM multi-head attention mechanism~\cite{vaswani2017attention} attends to the relevant information in the current turn for a specific domain-slot pair. In the process of slot-utterance matching, SUMBT utilises BERT’s~\citep{devlin2019bert} \texttt{[CLS]} sequence embedding to represent the semantic concepts in the model ontology. Instead of using the single \texttt{[CLS]} embedding, we make use of the sequence of embeddings for the domain-slot description. We choose to make this expansion, as approaches which utilise a sequence of embeddings outperform approaches based on a single embedding in various natural language processing tasks~\citep{poerner-etal-2020-sentence, choi2021evaluation}. We further use RoBERTa as a feature extractor~\cite{liu2019roberta}.

\paragraph{Dialogue context tracking} The first of the three components of the HIS model is a representation of the dialogue context (history). In the SUMBT approach, a gated-recurrent unit mechanism tracks the most important information during a dialogue. The resulting context conditioned representations for the domain-slot pairs contain the relevant information from the dialogue history. Similar to the alteration in the SUM component, we represent the dialogue context as a sequence of representations. This sequence, $\bm{C}_t^s$, represents the dialogue context for domain-slot pair $s$ across turns $1$ to $t$, while it dimension being independent of $t$. Besides the above modification, we add a further step where we reduce this sequence of context representations to a single representation $\hat{\mathbf{y}}_t^s$. We do this reduction using a learned convolutional pooler, which we call the \emph{Set Pooler}. See Appendix Section \ref{app:implementation} for more details regarding the implementation.

\paragraph{User goal prediction} The second component of the HIS model is the user goal. This is typically the only component that neural tracing models explicitly model as a set of domain-slot-value pairs. Here, we follow the matching network approach~\citep{vinyals2016matching} utilised by SUMBT, where the predictive distribution is based on the similarity between the dialogue context and the value candidates. To obtain the similarity between the dialogue context and a value candidate we make use of cosine similarity, $S_{\cos}\left( \cdot, \cdot \right)$. Based on these similarity scores, we produce a predictive distribution, Equation~\ref{eq:user_goal}, for the value of domain-slot pair $s$ at turn $t$ $v_t^s$, the user and system utterances at turn $t$, $\bm{u}_{t}^{usr}$ and $\bm{u}_{t-1}^{sys}$, and the dialogue context representations at turn $t-1$ $\bm{C}_{t-1}^s$. Contrary to the SUMBT approach, each value candidate is represented by the sequence of value description embeddings from a fixed RoBERTa model. The \emph{Set Pooler}, with the same parameters used for pooling context representations, reduces this sequence of value description representations to a representation $\mathbf{y}_v$, for value $v$.
\begin{equation}
\begin{split}
    {\tt P} \left(v_t^s = v | \bm{u}_{t}^{usr}, \bm{u}_{t-1}^{sys}, \bm{C}_{t-1}^s \right) = \\
    \frac{\exp \left( S_{\cos} \left( \hat{\mathbf{y}}_t^s, \mathbf{y}_v \right) \right)}{\sum_{v'} \exp \left( S_{\cos} \left( \hat{\mathbf{y}}_t^s, \mathbf{y}_{v'} \right) \right)},
\end{split}
\label{eq:user_goal}
\end{equation}

\paragraph{User action prediction} To be fully in line with the HIS model, we further require the predicted user actions. In order to predict the user actions, we categorise them into general user actions and user request actions. Further, since our system is a multi-domain system, we include the current active domain in the hidden information state of the system.

General user action includes actions such as the user thanking or greeting the system, which do not rely on the dialogue context. Hence, we can infer general user actions from the current user utterance. A user request action is an action indicating that the user is requesting information about an entity. \citet{zhu-etal-2020-convlab} shows that simple rule-based estimates of these actions lead to poor downstream policy performance. Hence, we propose predicting this information within the belief tracking model.

Since we can infer the general actions from the current user utterance, we use a single turn representation $\bm{x}_t^0$ to predict such actions. The single turn representation, $\bm{x}_t^0$, is the representation for the RoBERTa sequence representation \texttt{<s>}, which is equivalent to the BERT \texttt{[CLS]} representation. That is:
\begin{equation}
\begin{split}
    {\tt P} \left( a_t^{gen} = a | \mathbf{u}_{t}^{usr}, \mathbf{u}_{t-1}^{sys} \right)  = \\ \textrm{softmax} \left( \bm{W}^{gen} \mathbf{x}_t^0 + \bm{b}^{gen} \right),
    \label{nbt:gen}
\end{split}
\end{equation}
where $a \in \{ \texttt{none}, \texttt{thank\_you}, \texttt{goodbye} \}$.

The more difficult sub-tasks include active user request and active domain prediction. For user request prediction we utilise the dialogue context representation $\hat{\mathbf{y}}_t^s$ for a specific domain-slot pair to predict whether the user has requested information relating to this slot.
That is:
\begin{empheq}{align}
\begin{split}
    {\tt P} \left( r_t^s = 1 | \bm{u}_{t}^{usr}, \bm{u}_{t-1}^{sys}, \bm{C}_{t-1}^s \right) = \\ \textrm{sigmoid} \left( \bm{w}^{req} \hat{\mathbf{y}}_t^s + b^{req} \right),
    \label{nbt:request}
\end{split}
\end{empheq}
where $r_t^s$ indicates an active request for domain-slot $s$ by the user in turn $t$.

Last, to predict active domains in the dialogue, we incorporate information relating to all slots associated with a specific domain. We do so by performing mean reduction across the context representations of all the slots associated with a domain. The resulting domain representations are used to predict whether a domain is currently being discussed in the dialogue. That is, for active domain $d_t$, $S_d$ the set of slots within domain $d$, and $\bm{C}_{t-1}^d := \left[ \bm{C}_{t-1}^s \right]_{s \in S_d}$ the set of context representations for all domain-slot pairs in $S_d$ at turn $t-1$, we have the active domain distribution:
\begin{empheq}{align}
    \begin{split}
    {\tt P} \left( d_t = d | \bm{u}_{t}^{usr}, \bm{u}_{t-1}^{sys}, \bm{C}_{t-1}^d \right) = \\
    \textrm{sigmoid} \left( \bm{w}^{dom} \frac{1}{|S_d|}\sum_{s \in S_d} \hat{\mathbf{y}}_t^{s} + b^{dom} \right),
    \end{split}
    \label{nbt:domain}
\end{empheq}

\paragraph{Optimisation} For each of the four tasks: user goal prediction, general user action prediction, user request action prediction and active domain prediction, the aim of the model is to predict the correct class. To optimise for these objectives, we minimise the following classification loss functions: $\mathcal{L}_{\text{goal}}$, $\mathcal{L}_{\text{general}}$, $\mathcal{L}_{\text{request}}$ and $\mathcal{L}_{\text{domain}}$.
During model training we combine four weighted classification objectives:
\begin{empheq}{align}
    \begin{split}
    \mathcal{L} = \alpha_{\text{goal}} \mathcal{L}_{\text{goal}} + \alpha_{\text{general}} \mathcal{L}_{\text{general}} \\ + \alpha_{\text{request}} \mathcal{L}_{\text{request}} + \alpha_{\text{domain}} \mathcal{L}_{\text{domain}},
    \end{split}
\end{empheq}
where $\alpha_x \in \left( 0, 1 \right]$ is the importance of task $x$. In this work, we use the label smoothing classification loss for all sub-tasks as it results in better calibrated predictions, as shown by \citet{van-niekerk-2020-knowing}, see details in Section \ref{app:lsloss} of the appendix.

\subsection{Uncertainty Estimation in SetSUMBT}

Similarly to~\citet{van-niekerk-2020-knowing}, we construct an ensemble of SetSUMBT models by training each model on one of $10$ randomly selected subsets of data. We then distil this ensemble into a single model by adopting ensemble distillation (EnD) and ensemble distribution distillation (EnD$^2$) as described in Section~\ref{sec:distillation}. We refer to these distilled versions of the SetSUMBT ensemble as EnD-SetSUMBT and EnD$^2$-SetSUMBT, respectively. 

The SetSUMBT belief tracker tracks the presence and value of each domain-slot pair $s$ as the dialogue progresses. For the sake of scalability of the downstream policy, in the user goal $g$ we do not consider all possible values, but rather the most likely one $v^s$ for every domain-slot pair $s$ and its associated probability, i.e., the confidence score given by $h_{t,s}^{g}$ summarised in vector $\mathbf{h}_{t}^{g}$ for all domain-slot pairs:
%
\begin{align}
    v^s &\coloneqq \arg\max_{v} {\tt P} \left(v_t^s = v | \bm{u}_{t}^{usr}, \bm{u}_{t-1}^{sys}, \bm{C}_{t-1}^s \right), \notag \\
    h_{t,s}^{g} &\coloneqq \max_{v} {\tt P} \left(v_t^s = v | \bm{u}_{t}^{usr}, \bm{u}_{t-1}^{sys}, \bm{C}_{t-1}^s \right),\notag \\
    \mathbf{h}_{t}^{g} &\coloneqq [v^s,h_{t,s}^{g}]_{\forall s}.
    \label{bs:goal}
\end{align}
For the EnD-SetSUMBT belief tracker, we can also calculate the total uncertainty for each domain-slot given by the entropy, see Section~\ref{sec:distillation}. We encode that information  in $h_{t,s}^{unc}$ for each domain-slot pair $s$ and summarise in $\mathbf{h}_{t}^{unc}$ for all domain-slot pairs:
\begin{align*}
     h_{t,s}^{unc} &\coloneqq \mathcal{H} \left[ {\tt P} \left(v_t^s = v | \bm{u}_{t}^{usr}, \bm{u}_{t-1}^{sys}, \bm{C}_{t-1}^s \right) \right], \\
     \mathbf{h}_{t}^{unc} &\coloneqq  [h_{t,s}^{unc}]_{\forall s}.
\end{align*}
For the EnD$^{2}$-SetSUMBT belief tracker, can further include the knowledge uncertainty for each domain-slot pair $s$ given by the mutual information:
\begin{equation*}
     h_{t,s}^{unc} \coloneqq \mathcal{I}[v_t^s,{\tt \bm{\pi}}|\bm{u}_{t}^{usr}, \bm{u}_{t-1}^{sys}, \bm{C}_{t-1}^s;\bm{\phi}],
\end{equation*}
as per Eq.~\eqref{eqn:mipn} where $\bm{\pi}$ represents the ensemble distribution and $\bm{\phi}$ the model parameters.

In addition, all versions of SetSUMBT include the following vectors/variables:

\begin{compactdesc}
\item [$\mathbf{h}_{t}^{g}$] is the estimate of the user goal from Eq.~\eqref{bs:goal},
\item [$\mathbf{h}_t^{usr}$] is the estimate of user actions from Eq.~(\ref{nbt:gen}-\ref{nbt:domain}),
\item [$\mathbf{h}_t^{db}$] is the database search result\footnote{Uncertainty is incorporated in the database query vector in the form of confidence thresholds. If the confidence score for a specific constraint is less than a chance prediction then this constraint is ignored during the database query.},
\item [$\mathbf{h}_{t-1}^{sys}$] is the system action,
\item [$\mathbf{h}_t^{book}$] is the set of completed bookings,
\item [$h_t^{term}$]indicates the termination of the dialogue.
\end{compactdesc}

This results in the following belief state:
\begin{equation*}
    \mathbf{b}_t = \{ \mathbf{h}_{t}^{usr}, \mathbf{h}_{t-1}^{sys}, \mathbf{h}_{t}^{g}, \mathbf{h}_t^{book}, \mathbf{h}_t^{db}, h_t^{term},
    \mathbf{h}_{t}^{unc}\}.
\end{equation*}

For a system without uncertainty, all confidences would be rounded to either $0$ or $1$ and the belief state would not contain the $\mathbf{h}_{t}^{unc}$ vector.

\subsection{Policy Optimisation as Downstream Task}\label{sec:downstream_policy}

For our experiments we optimise the dialogue policy operating on the belief state via RL using the PPO algorithm \cite{ppo}. PPO is an on-policy actor-critic algorithm that is widely applied across different reinforcement learning tasks because of its good performance and simplicity. Similarly to~\citet{gdpl}, we use supervised learning to pretrain the policy before starting the RL training phase. In order to perform supervised learning we need to map the belief states into system actions as they occur in the corpus. These belief states can either be oracle states taken from the corpus or predictions of our belief tracker that takes corpus dialogues as input. We investigate both options for policy training. 


\section{Experiments}

\subsection{Neural Belief Tracking Performance}

\paragraph{Overall performance} Table~\ref{tab:nbt} compares the performance of our proposed SetSUMBT belief tracker to existing approaches, which include SUMBT, the calibrated ensemble belief state tracker (CE-BST)~\citep{van-niekerk-2020-knowing} and the end to end trained SUMBT+LaRL approach~\citep{lee2020sumbt+}. We consider the joint goal accuracy (JGA), L2-Error and expected calibration error (ECE). The JGA of a belief tracking model is the percentage of turns for which the model correctly predicted the value for all domain-slot pairs. The L2-Error is the L2-Norm of the difference between the predicted user distribution and the true user goal. Further, the ECE is the average absolute difference between the accuracy and the confidence of a model. In this comparison, we do not consider state tracking approaches, as they do not yield uncertainty estimates. SetSUMBT outperforms SUMBT and SUMBT+LaRL in terms of calibration and accuracy. We name the variants of SetSUMBT as follows: CE-SetSUMBT is a calibrated ensemble of SetSUMBT similar to CE-BST, EnD-SetSUMBT is the distilled SetSUMBT model, and EnD$^{2}$-SetSUMBT is the distribution distilled SetSUMBT model.

\begin{table}[t]
    \centering
    \small
    \begin{tabular}{p{2.4cm}|p{1.2cm}|p{1.2cm}|p{1.2cm}}
        Approach & JGA(\%) & L2-Error & ECE(\%) \\
        \hline
        SUMBT & 46.78 & 1.1075 & 25.46 \\
        CE-BST & 48.71 & \textbf{1.1042} & 10.73 \\
        SUMBT+LaRL & 51.52 & - & - \\
        \hline
        SetSUMBT & 51.11 & 1.2386 & 15.13 \\
        EnD$^2$-SetSUMBT & 51.22 & 1.1948 & 7.09 \\
        CE-SetSUMBT & 52.04 & 1.1936 & \textbf{6.84} \\
        EnD-SetSUMBT & \textbf{52.26} & 1.1782 & 7.54 \\
        \hline
    \end{tabular}
    \caption{Comparison of neural belief tracking approaches on the MultiWOZ 2.1 test set. CE is an ensemble of calibrated models, EnD is ensemble distillation and EnD$^2$ is ensemble distribution distillation.}
    \label{tab:nbt}
\end{table}
\paragraph{Runtime efficiency} The single instance of the SetSUMBT tracker processes a dialogue turn in approximately $77.768$ ms, whereas an ensemble of $10$ models processes a turn in approximately $768.025$ ms. These processing times are averaged across the $7372$ turns in the MultiWOZ test set, see Appendix Section \ref{app:latency} for more details. The significant increase in processing time for the ensemble of models makes this approach inappropriate for real time interaction with users on a private device.
%
%
\begin{figure}
    \centering
    \includegraphics[scale=0.25]{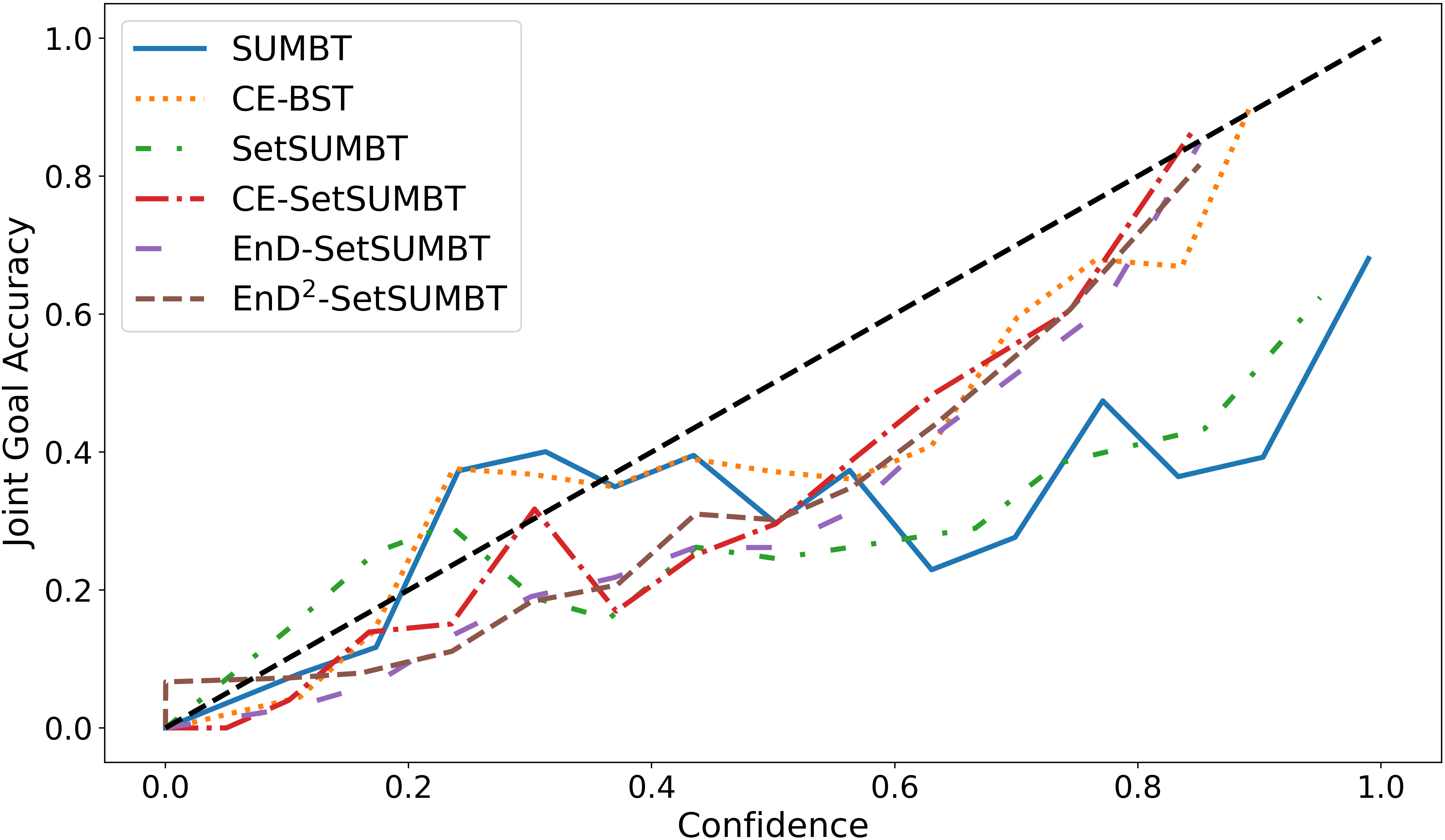}
    \caption{Reliability comparison of a selection of neural belief tracking models. 
    }
    \label{fig:reliability}
    \vspace{-10pt}
\end{figure}
\paragraph{Calibration} The reliability diagram in Figure~\ref{fig:reliability} illustrates the relationship between the joint goal accuracy and the model confidence. The best calibrated model is the one that is closest to the diagonal, i.e., the one whose confidence for each dialogue state is closest to the achieved accuracy. The best reliability is achieved by CE-BST, and CE-SetSUMBT comes second. Both distillation models (EnD-SetSUMBT and EnD$^2$-SetSUMBT) do not deviate greatly from CE-SetSUMBT.
%
\subsection{Policy Training on User Simulator}\label{sec:simulated}

We incorporate SetSUMBT, EnD-SetSUMBT and EnD$^2$-SetSUMBT within the Convlab2 \cite{zhu-etal-2020-convlab} task-oriented dialogue environment and compare their performance by training policies which take belief states as inputs\footnote{\url{https://gitlab.cs.uni-duesseldorf.de/general/dsml/setsumbt-public.git}}.

To investigate the impact of additional uncertainty measures on the dialogue policy we perform interactive learning in a more challenging environment than the original Convlab2 template-based simulator. We add ambiguity to the simulated user utterances in the form of value variations that occur in the MultiWOZ dataset. For example, instead of the user simulator asking for a hotel for "one person", it could also say "It will be just me.". For more information see Appendix Section~\ref{app:us_noise}.

When policies are trained for large domains, they are typically first pretrained on the corpus in a supervised manner, and then improved using reinforcement learning. We first investigate which states to use for the supervised pretraining
(Section~\ref{sec:downstream_policy}): oracle states, i.e., the dialogue state labels from the MultiWOZ corpus, or estimated belief states, e.g., those predicted by a EnD-SetSUMBT model. We then evaluate the pretrained policies with the simulated user. During the evaluation both policies use a EnD-SetSUMBT model to provide belief states. We observe that the policy pretrained using the oracle state achieves a success rate of $36.50\%$ in the simulated environment compared to the $46.08\%$ success rate achieved by the policy pretrained using EnD-SetSUMBT. Thus, all our following experiments use predicted belief states of respective tracking models for the pretraining stage.

%
For each setting of the belief tracker we have four possible belief state settings, i.e., the binary state (no uncertainty), the confidence score state, the confidence score state with additional total uncertainty features and the confidence score state with additional knowledge uncertainty features. For each setting we evaluate the policies through interaction with the user simulator, results are given in Table~\ref{tab:simulated}. 

In interaction with the simulator, systems making use of confidence outperform the systems without any uncertainty (significance at $p<0.05$). Moreover, the additional total and knowledge uncertainty features always outperform the systems which only use a confidence score (significance at $p<0.05$). This indicates that additional measures of uncertainty improve the robustness of the downstream dialogue policy in a challenging environment.

It is interesting to note that the system which makes use of total uncertainty appears to outperform the system that makes use of knowledge uncertainty (significance at $p<0.05$). We suspect that this controlled simulated environment has low data uncertainty, so the total uncertainty is overall more informative.
\begin{table}[t]
    \centering
    \small
    \begin{tabular}{p{1.3cm}p{1.5cm}|x{1.05cm}x{.85cm}x{.85cm}}
         \multirow{2}{1.4cm}{Belief Tracker} & Belief state uncertainty & Success Rate & Reward & Turns \\
         \hline
         \multirow{2}{*}{SetSUMBT} & None & 78.67 & 46.51 & 7.49 \\
         & Confidence & 83.25 & 52.49 & 6.80 \\
         \hline
         \multirow{3}{1.4cm}{EnD-SetSUMBT} & None & 82.25 & 51.18 & 7.52 \\
         & Confidence & 83.75 & 54.04 & \textbf{6.46} \\
         & Total & \textbf{86.83} & \textbf{57.09} & 7.35 \\
         \hline
         \multirow{4}{1.4cm}{EnD$^2$-SetSUMBT} & None & 83.75 & 53.00 & 7.50 \\
         & Confidence & 84.08 & 53.15 & 7.74 \\
         & Total & 84.83 & 54.54 & 7.26 \\
         & Knowledge & 85.17 & 54.63 & 7.57 \\
         \hline
    \end{tabular}
    \caption{Performance of the systems in the simulated environment. For each setting we have $5$ policies initiated with different random seeds, each evaluated with $1000$ dialogues and their success rates, reward and number of turns averaged.}
    \label{tab:simulated}
    \vspace{-10pt}
\end{table}
\subsection{Human Trial}

We conduct a human trial, where we compare SetSUMBT as the baseline with EnD-SetSUMBT and EnD$^2$-SetSUMBT. For EnD-SetSUMBT, we consider the model that includes both confidence scores and entropy features. For EnD$^2$-SetSUMBT, we investigate the model that includes confidence scores and knowledge uncertainty features. For each model we have two variations: one with a binary state corresponding to the most likely state (no uncertainty variation), and one with uncertainty measures (uncertainty variation). For each variation we chose the policy whose performance on the simulated user is closest to the average performance of its respective setting, see Section~\ref{sec:simulated}.

Subjects are recruited through the Amazon Mechanical Turk platform to interact with our systems via the DialCrowd platform~\cite{lee-etal-2018-dialcrowd}. Each assignment consists of a dialogue task and two dialogues to perform. The task comprises a set of constraints and goals, for example finding the name and phone number of a guest house in the downtown area. We encourage the subjects to use variants of labels by introducing random value variants in the tasks. The two dialogues are performed in a random order with two variations of the same model, namely no-uncertainty and uncertainty variation, as described above. After each dialogue, the subject rates the system as successful if they think they received all the information required and all constraints were met. The subjects rate each system on a 5 point Likert scale. In total we collected approximately $550$ dialogues for each of $6$ different systems, $3300$ in total. There was a total of $380$ subjects who took part in these experiments.

Table~\ref{tab:trial} shows the performance of the above policies in the human trial. We confirm that each no uncertainty system is always worse than its uncertainty counterpart (each significant at $p<0.05$). It is important to emphasise here that in each pairing, the systems have exactly the same JGA, but their final performance can be very different in terms of success and user rating. This empirically demonstrates the limitations of JGA as a single measure for dialogue state tracking, urging the modelling of uncertainty and utilisation of calibration measures.
Finally, we observe that adding additional uncertainty measures improves the policy (each significant at $p<0.05$) and the best overall performance is achieved by the system that utilises both knowledge uncertainty and confidence scores (significant at $p<0.05$). This suggests that in human interaction there is more data uncertainty, necessitating the knowledge uncertainty to be an explicit part of the model.

It is important to note here that solely a lower average number of turns is not necessarily an indicator of the desired behaviour of a system. For example, a system which says goodbye too early may never be successful, but will have a low average number of turns.
%
\begin{table}[t]
    \centering
    \small
    \begin{tabular}{p{1.3cm}p{1.5cm}|x{1.05cm}x{.85cm}x{.85cm}}
        \multirow{2}{1.3cm}{Belief Tracker} &  Belief state uncertainty & Success Rate & Rating & Turns \\
        \hline
        \multirow{2}{*}{SetSUMBT} & None & 48.99 & 2.68 & 7.28 \\
        & Confidence & 67.05 & 3.47 & \textbf{6.12} \\
        \hline
        \multirow{2}{1.3cm}{EnD-SetSUMBT} & None & 64.09 & 3.29 & 6.45 \\
        & Total & 68.25 & 3.36 & 6.45 \\
        \hline
        \multirow{2}{1.3cm}{EnD$^2$-SetSUMBT} & None & 66.25 & 3.35 & 6.25 \\
        & Knowledge & \textbf{71.61} & \textbf{3.52} & 6.31 \\
        \hline
    \end{tabular}
    \caption{Performance of the systems evaluated with real users. We have $550$ dialogues for each system with success rates, ratings and the number of turns averaged. }
    \label{tab:trial}
    \vspace{-10pt}
\end{table}
\section{Conclusion}

Whilst neural dialogue state trackers may achieve state-of-the-art performance in the isolated dialogue state tracking task, the absence of uncertainty estimates may lead to less robust performance of the downstream dialogue policy. In this work we propose the use of total and knowledge uncertainties along with confidence scores to form a dialogue belief state. We moreover describe a model, SetSUMBT, that can produce such a belief state via distillation. Experiments with both simulated and real users confirm that these uncertainty metrics can lead to more robust dialogue policy models. In future, we will investigate modifying span-based dialogue state trackers to incorporate uncertainty. We will further investigate the expansion of the SetSUMBT model to include the correlation between different domain-slot pairings.

\section*{Acknowledgements}

CVN, MH, NL and SF are supported by funding provided by the Alexander von Humboldt Foundation in the framework of the Sofja Kovalevskaja Award endowed by the Federal Ministry of Education and Research. CG and HCL are supported by funds from the European Research Council (ERC) provided under the Horizon 2020 research and innovation programme (Grant agreement No. STG2018 804636). Google Cloud and HHU ZIM provided computational infrastructure.

\bibliography{refs}
\bibliographystyle{acl_natbib}

\appendix
\section{Example dialogues}
Here we provide two example dialogues from our human trial, which illustrates the advantage of using uncertainty. In Figures~\ref{fig:dial_eg} and \ref{fig:dial_eg1} we observe that the system makes a second request to the user to clarify the intent of the user.
\begin{figure}[ht]
    \fbox{\begin{minipage}{19em}
    \vspace{0.1cm}
    User: I need to book a cheap 4 star hotel.

    System: Do you have a \textbf{area} preference?

    User: I need the hotel to be \textbf{eastern}.

    System: Is there a specific \textbf{area} of town you are interested in?

    User: The \textbf{east}.
    \vspace{0.1cm}
    \end{minipage}}
    \caption{Extract from a dialogue with a real user illustrating the importance of uncertainty in a dialogue.}
    \label{fig:dial_eg}
\end{figure}
\vspace{-0.6cm}
\begin{figure}[ht]
    \fbox{\begin{minipage}{19em}
    \vspace{0.1cm}
    User: Find me a hotel in the city centre.

    System: Do you have a price range preference?

    User: \textbf{High}

    System: It is in the centre. \textbf{What price range would you like?}

    User: A \textbf{highly priced} hotel.

    System: Everyone seems to enjoy Acorn guest house. \textbf{It is an expensive hotel}. Is there anything else I can help you with?
    \vspace{0.1cm}
    \end{minipage}}
    \caption{Extract from a dialogue with a real user illustrating the importance of uncertainty in a dialogue.}
    \label{fig:dial_eg1}
\end{figure}

\section{Loss Functions}

Consider the classification problem with input features $\bm{x}$, oracle class $y$ and training dataset: \begin{empheq}{align*}
\mathcal{D}_{\text{train}} = \{ (\bm{x}^{(1)}, y^{(1)}), ..., (\bm{x}^{(N)}, y^{(N)}) \},
\end{empheq}
consisting of $N$ labelled examples.
\subsection{Label Smoothing Loss Function}\label{app:lsloss}
The label smoothing loss is a regularised variant of standard negative log likelihood loss. Here, instead of considering a 1-hot target $\bm{y}_I$, some noise is induced in the target distribution in the form of:
\begin{empheq}{align*}
    \bm{y}_{ls} = (1 - \epsilon) \bm{y}_I + \frac{\epsilon}{K},
\end{empheq}
where $\epsilon$ is the smoothing parameter, $\bm{y}_{ls}$ the noisy/smoothed targets and $\bm{y}_I$ the one hot representation of the target $y$. The objective is to minimise the KL divergence between the predictive distribution, ${\tt P}(y|\bm{x}^{(i)}, \bm{\phi})$, and the smoothed target $\bm{y}_{ls}$. That is
\begin{empheq}{align*}
    \mathcal{L}_{ls} (\bm{\phi}, \mathcal{D}_{\text{train}}) = \frac{1}{N} \sum_{i=1}^N {\tt KL} \big[ \bm{y}_{ls}^{(i)}  || {\tt P}(y|\bm{x}^{(i)}, \bm{\phi}) \big]
\end{empheq}

\subsection{Distillation Loss Functions}\label{app:end2loss}

Here we detail the loss functions used for ensemble distillation (EnD) and ensemble distribution distillation (EnD$^2$) in this work.

Consider an ensemble $\{ \bm{\theta}^{(1)}, ..., \bm{\theta}^{(M)} \}$ consisting of $M$ models, with predictive posterior ${\tt P}(y|\bm{x}^{(i)}, \mathcal{D}_{\text{train}})$.

Standard ensemble distillation~\cite{hinton2015distilling} is accomplished by minimising the KL-divergence between a student model with parameters $\bm{\phi}$ and the ensemble's predictive posterior:
\begin{empheq}{align*}
\begin{split}
&\mathcal{L}_{\text{EnD}}(\bm{\phi},\mathcal{D}_{\text{train}}) = \\
& \ \ \frac{1}{N}\sum_{i=1}^N {\tt KL}\big[{\tt P}(y|\bm{x}^{(i)}, \mathcal{D}_{\text{train}})\ || \  {\tt P}(y|\bm{x}^{(i)}, \bm{\phi})\big]
\end{split}
\label{eqn:end-loss}
\end{empheq}

Distribution distillation is accomplished using the improved loss function proposed by \citet{malinin2021send}. Here, we first compute a \emph{Proxy Dirichlet Target} with Dirichlet concentration parameters $\bm{\beta}$ from the ensemble:
\begin{empheq}{align}
\begin{split}
    &    \hat \pi_k (\bm{x})\small{=} \frac{1}{M}\sum_{m=1}^M {\tt P}(y=\omega_k|\bm{x}, \bm{\theta}^{(m)}) \\
     &   \tilde \beta_0(\bm{x}) \small{=} \frac{K-1}{2 \sum_{k=1}^K\hat \pi_k (\ln \hat \pi_k \small{-}\sum_{m=1}^M\frac{\ln \pi_k^{(m)}}{M})} \\
     &   \beta_k (\bm{x}) \small{=} \hat \pi_k(\bm{x}) \cdot \tilde \beta_0(\bm{x}) + 1,\ \beta_0 = \sum_{k=1}^K \beta_k.
\end{split}
\end{empheq}
Given this \emph{Proxy Dirichlet Target}, distribution distillation is done by minimising the following loss:
\begin{empheq}{align}
\begin{split}
&\mathcal{L}_{\text{EnD}^2}(\bm{\phi},\mathcal{D}_{\text{train}}) = \\
 &\ \ \frac{1}{N}\sum_{i=1}^N\Big[\small{-}\mathbb{E}_{{\tt p}(\bm{\pi}|\bm{x}^{(i)},\bm{\phi})}[\small\sum_{k=1}^K\hat \pi_k^{(i)}\ln \pi_k] \\
&\ \ +\frac{1}{\beta_0^{(i)}}{\tt KL}[{\tt p}(\bm{\pi}|\bm{x}^{(i)},\bm{\phi}) \| {\tt p}(\bm{\pi}|\bm{1})]\Big].
\end{split}
\label{eqn:endd-loss}
\end{empheq}

\section{SetSUMBT Implementation Details}
\label{app:implementation}
Here we provide details regarding the SetSUMBT model configuration and the model training configuration. Table \ref{tab:setsumbt_config} provides details about the configuration of the SetSUMBT model. Tables \ref{tab:train_config} and \ref{tab:dist_train_config} provide details regarding the training configurations for both the single model and distillation (EnD) of SetSUMBT. For all SetSUMBT models the \emph{Set Pooler} consists of a single convolutional layer with padding followed by a mean pooling layer.
\begin{table}[ht]
    \small
    \centering
    \begin{tabular}{c|c}
        Parameter & Value \\
        \hline
        Roberta pretrained checkpoint & \texttt{roberta-base} \\
        Hidden size & $768$ \\
        SUM attention heads & $12$ \\
        Context tracking GRU hidden size & $300$ \\
        \emph{Set Pooler} CNN filter size & $3$ \\
        Dropout rate & $0.3$ \\
        Maximum turn length & $64$ \\
        Candidate description length & $12$
    \end{tabular}
    \caption{SetSUMBT model configuration.}
    \label{tab:setsumbt_config}
\end{table}
\begin{table}[ht]
    \small
    \centering
    \begin{tabular}{c|c}
        Parameter & Value \\
        \hline
        Learning rate (LR) & $1e-5$ \\
        LR Scheduler warmup proportion & $0.1$ \\
        Batch size & $3$ \\
        Maximum turns per dialogue & $12$ \\
        Epochs & $100$ \\
        Early stopping criteria & $25$ epochs \\
        Label smoothing $\epsilon$ & $0.05$ \\
        $\alpha_{\text{goal}}$ & 1.0 \\
        $\alpha_{\text{general}}$ & 0.2 \\
        $\alpha_{\text{request}}$ & 0.2 \\
        $\alpha_{\text{domain}}$ & 0.2
    \end{tabular}
    \caption{Single model and EnD$^2$ training configurations. EnD$^2$ does utilise the Label smoothing $\epsilon$.}
    \label{tab:train_config}
\end{table}
\begin{table}[ht]
    \small
    \centering
    \begin{tabular}{c|c}
        Parameter & Value \\
        \hline
        Learning rate (LR) & $1e-5$ \\
        LR Scheduler warmup proportion & $0.1$ \\
        Batch size & $3$ \\
        Maximum turns per dialogue & $12$ \\
        Epochs & $100$ \\
        Early stopping criteria & $25$ epochs \\
        Distribution smoothing & $1e-4$ \\
        Temperature scaling base temperature & $2.5$ \\
        Temperature scaling annealing cycle & $0.1$ \\
        $\alpha_{\text{goal}}$ & 1.0 \\
        $\alpha_{\text{general}}$ & 0.2 \\
        $\alpha_{\text{request}}$ & 0.2 \\
        $\alpha_{\text{domain}}$ & 0.2
    \end{tabular}
    \caption{EnD training configuration.}
    \label{tab:dist_train_config}
\end{table}

\section{Variations in User Simulator Output}
\label{app:us_noise}

The user simulator used in our experiments consists of a natural language understanding (NLU) module, a rule based user agent and template based natural language generation module, all provided in the ConvLab $2$ environment~\citep{zhu-etal-2020-convlab}. A pre-defined set of rules simulates the user behaviour based on the predicted semantic system actions and the resulting user actions are mapped to natural language using a pre-defined set of templates. To induce variation to the user simulator utterances and thus make understanding more difficult for the system, we utilise a set of pre-defined value variations obtained from the MultiWOZ $2.1$ value map ~\citep{heck2020trippy}. For example, we can map the value, \texttt{expensive}, in the user action:
\begin{center}
\texttt{Inform - Restaurant - Price\_range - expensive}
\end{center}
to any of the following options:
\begin{center}
[ \texttt{high end, high class, high scale, high price, high priced, higher price, fancy, upscale, nice, expensively, luxury}
].
\end{center}
In our experiments $20\%$ of simulated user actions contain such variations. 

\section{System Latencies}
\label{app:latency}

In this section we provide the processing times per turn for our SetSUMBT model as well as the systems used in this work. These processing times are averaged across the $7372$ turns in the MultiWOZ 2.1 test set. This test is performed on a Google Cloud virtual machine containing a Nvidia V100 16GB GPU, 8 n1-standard VCPU’s and 30GB memory. In Table \ref{tab:setsumbt_latencies} we compare the latencies of a single instance of SetSUMBT against a $10$ model ensemble. In Table \ref{tab:sys_latencies} we compare the latencies of the full dialogue system setups used in this work.
\begin{table}[ht]
    \centering
    \begin{tabular}{c|c}
        Tracker & Latency (ms) \\
        \hline
        Single instance & $77.7680$ \\
        10 Instance ensemble & $768.0256$ \\
        \hline
    \end{tabular}
    \caption{Turn level latency of the SetSUMBT model and ensemble.}
    \label{tab:setsumbt_latencies}
\end{table}
\vspace{-0.5cm}
\begin{table}[ht]
    \centering
    \begin{tabular}{c|c}
        System & Latency (ms) \\
        \hline
        No uncertainty & $135.9768$ \\
        Confidence scores & $138.0960$ \\
        Total uncertainty & $147.4574$ \\
        Knowledge uncertainty & $152.5392$ \\
        \hline
    \end{tabular}
    \caption{Turn level latency of the various full dialogue systems utilised in this work.}
    \label{tab:sys_latencies}
\end{table}
\end{document}